\documentclass[]{article}
\usepackage{xcolor}
\usepackage{spconf,amsmath,graphicx}
\usepackage{graphicx}
\usepackage{amsmath}
\usepackage{amssymb}
\usepackage{booktabs}
\usepackage{multirow}
\usepackage{graphicx} 
\usepackage{epstopdf}
\usepackage{tabularx}
\usepackage{xcolor}
\usepackage{amssymb}
\usepackage{colortbl}
\usepackage{hyperref}

\definecolor{grey}{RGB}{127,127,127}

\title{Differentiable Resolution Compression and Alignment for Efficient Video Classification and Retrieval}  

\name{Rui Deng\textsuperscript{1}, Qian Wu\textsuperscript{1}, Yuke Li\textsuperscript{1}, Haoran Fu\textsuperscript{2}}
\address{\textsuperscript{1}NetEase Yidun AI Lab    \qquad  \textsuperscript{2}Zhejiang University }
\begin{document}
\ninept
\maketitle
\begin{abstract}
Optimizing video inference efficiency has become increasingly important with the growing demand for video analysis in various fields. Some existing methods achieve high efficiency by explicit discard of spatial or temporal information, which poses challenges in fast-changing and fine-grained scenarios. To address these issues, we propose an efficient video representation network with \textbf{Differentiable Resolution Compression and Alignment} mechanism, which compresses non-essential information in the early stage of the network to reduce computational costs while maintaining consistent temporal correlations. Specifically, we leverage a \textbf{Differentiable Context-aware Compression Module} to encode the saliency and non-saliency frame features, refining and updating the features into a high-low resolution video sequence. To process the new sequence, we introduce a new \textbf{Resolution-Align Transformer Layer} to capture global temporal correlations among frame features with different resolutions, while reducing spatial computation costs quadratically by utilizing fewer spatial tokens in low-resolution non-saliency frames. The entire network can be end-to-end optimized via the integration of the differentiable compression module. Experimental results show that our method achieves the best trade-off between efficiency and performance on near-duplicate video retrieval and competitive results on dynamic video classification compared to state-of-the-art methods.\textit{Code:\href{https://github.com/dun-research/DRCA}{\color{blue}https://github.com/dun-research/DRCA}}

\end{abstract}
\vspace{-0.5em}
\section{Introduction}
\label{sec:intro}
\vspace{-0.5em}

Video representation learning is a crucial research topic due to its numerous applications, such as recommendation system, video search, etc. Due to the high redundancy across video frames, a judicious and effective strategy to expedite video inference involves mitigating redundant computations. Some existing \textbf{\textit{skipping-based}} methods\cite{gowda2021smart, adafocusplus, ocsampler, xia2022temporal, raviv2022dstep} propose to skip non-saliency frames to save computation costs via various temporal saliency sampling strategies. These methods are based on the assumption that the most saliency frame/regions contribute the most to the video representation. However, due to the inevitable loss of spatial information and inaccurate saliency measures, these methods encounter challenges in some fast-changing and fine-grained scenarios. For example, discriminating between different dancing types is concealed within the gradual transitions of each frame. Besides, for tasks that rely heavily on detailed information, such as video retrieval, discarding non-saliency frames may result in losing essential clues for accurate matching.
In contrast, some \textbf{\textit{multi-network-based}} methods\cite{liteeval, arnet, sun2021dynamic} have been proposed to retain non-saliency frames while using multiple networks with different computational costs to mitigate the overall expenses. Although fusion layers are designed to integrate multiple streams of information, extracting global temporal information may be inadequate in representing fine-grained information to achieve precise contextual understanding.
Besides, several of the above algorithms select different numbers of saliency frames based on the video content, making batched inference complicated and challenging.

\begin{figure*}[t]
  \centering
  \includegraphics[width=0.90\linewidth]{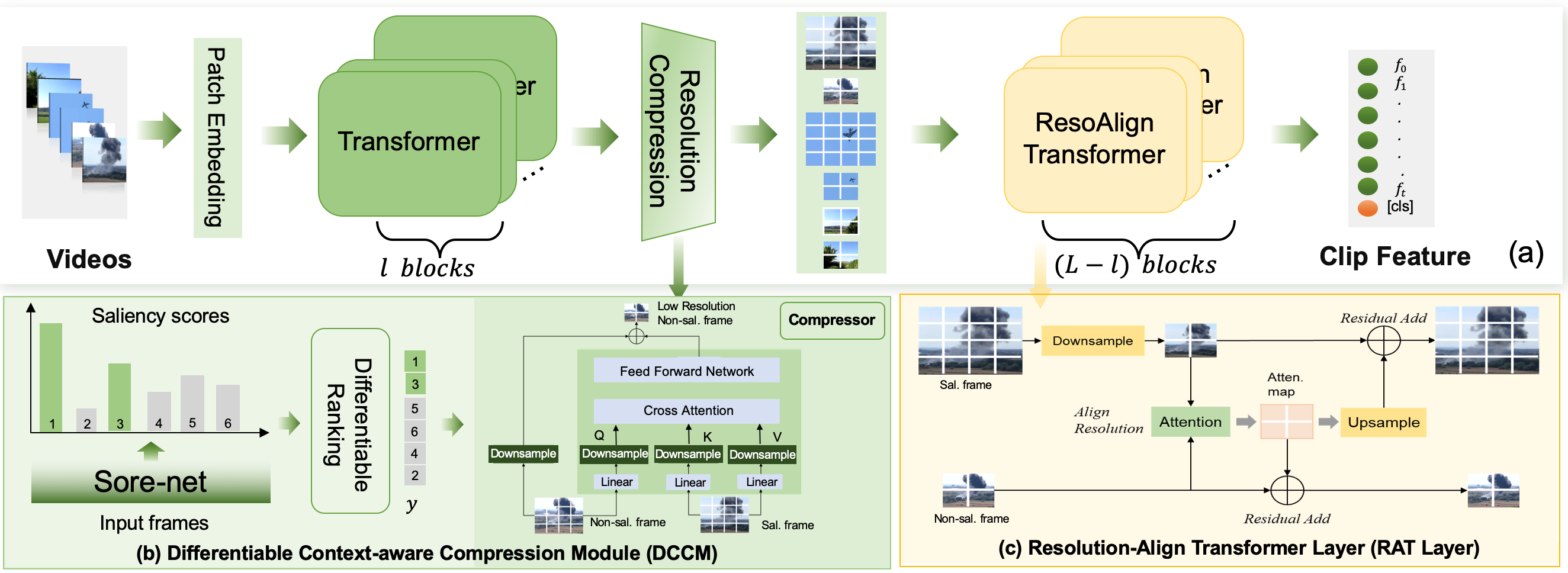}
  \caption{(a) The Overview of DRCA. (b) The frame sequence fed into DCCM is divided into saliency and non-saliency frames based on the saliency scores. Then the non-saliency frames are compressed with saliency frames as a reference to minimize information loss. In the training stage, the score-net can be updated through differentiable ranking by solving the optimal sorted indices $y$.(c) The compressed frame sequence with different resolutions is fed into RAT Layers to extract features and compute the final classification score or retrieval embedding.   }
  \vspace{-1.5em}
  \label{fig:overview}
\end{figure*}



To address these issues, in this paper, we cast efficient video learning as a Differentiable Resolution Compression and Alignment \textbf{(DRCA)} task, which consists of two main components to accomplish resolution compression and alignment. We instantiate the two components as \textbf{DCCM} (Differentiable Context-aware Compression Module) and \textbf{RAT-Layer} (Resolution-Align Transformer Layer).


Specifically, to reduce video tokens and minimize information loss, unlike previous methods\cite{arnet, adafocusplus, sun2021dynamic} that use an additional network to measure frame saliency separately, our \textbf{DCCM} incorporates the saliency assessment and frame compression directly into the network, enabling them to be not only context-aware but also learnable. Concretely, the DCCM can be end-to-end optimized through differentiable ranking, which solves an optimal saliency sorting sequence based on the predicted scores. By reducing the resolution of non-salient frames and decreasing the number of spatial tokens, significant reductions in spatial computational costs can be achieved, as the computational complexity of the Transformer increases quadratically for the number of tokens. Meanwhile, this also poses a challenge for existing networks like 3D convolutional neural networks\cite{feichtenhofer2019slowfast, lin2019tsm} and Transformers\cite{timesformer, mvit, swin}, as they lack the capability to directly extract features in this multi-resolution context. Therefore, we propose a simple but effective \textbf{RAT Layer} to process the frame tokens with different resolutions. The RAT Layer can extract strong spatial-temporal correlations layer-by-layer, which is crucial for video tasks that require reasoning, such as retrieving the same incident, where some video frames captured by different photographers are visually similar but not identical.


Notably, this novel approach enables efficient and accurate video representation learning while preserving the ability of batched inference with fixed token numbers. This is achieved by fully end-to-end optimizing integrated with a differentiable compression module. As a result, our method achieves the best trade-off between efficiency and performance compared to state-of-the-art (SOTA) methods in near-duplicated video retrieval(NDVR). Additionally, it demonstrates competitive performance in dynamic video classification compared to other SOTA methods.


Our contributions are summarized as follows:(1) a novel \textbf{Differentiable Resolution Compression and Alignment} network for efficient video learning(2) a novel \textbf{DCCM} designed to compress the non-essential information with differentiable ranking(3) a simple but effect \textbf{RAT Layer} to extract spatial-temporal correlations in multi-resolution compressed video sequence(4) achieving the best trade-off between efficiency and performance on NDVR task and competitive result on dynamic video classification task against SOTA.


\vspace{-1.5em}
\section{Related Works}
\vspace{-0.5em}
\noindent\textbf{Dynamic Video Inference} These approaches dynamically allocate computing resources based on input video to reduce overall computation costs. Some \textit{skipping-based} works adaptively skip non-saliency frames to reduce computational resource\cite{ gowda2021smart, adafocusplus, ocsampler, xia2022temporal, raviv2022dstep}.
However, this method completely discards the information on non-saliency frames, which results in a high dependence on the accuracy of saliency measures.
Conversely,  other \textit{multi-network-based}\cite{liteeval, arnet, sun2021dynamic} methods
propose to use lower-resource networks to extract information from
the non-saliency frames, while it may limit the model to extracting the global temporal information in the video sequences. Generally, these methods require some techniques to obtain a dynamic temporal strategy that determines how each frame should be processed, such as keeping, skipping, or other operations. 
Some methods\cite{gowda2021smart, adafocusplus, ocsampler, xia2022temporal,liteeval} utilize Reinforcement Learning\cite{rao2000reinforcement} to optimize the temporal process, but they often suffer from slow convergence. 
Other methods\cite{ raviv2022dstep, arnet, sun2021dynamic} use the Gumbel-Softmax Sampling trick\cite{gumblesoftmax} to optimize a temporal policy network, but it cannot explicitly control the number of selected saliency frames.

%

\vspace{-1em}
\section{Proposed Method}
\vspace{-0.5em}

The overview of our DRCA is shown in Fig \ref{fig:overview}. Given a video $V \in \mathbb{R}^{T \times H \times W \times 3}$ with T frames, our goal is to learn efficient and accurate features with fewer computational resources through adaptive compression in Visual Transformer\cite{vit}. Our DRCA achieves this goal by using two modules: DCCM and RAT layer, which are applied for resolution compression and alignment, respectively.


\begin{table*}[t]
\centering
\tabcolsep=9pt
\scalebox{0.85}{%

\begin{tabular}{c|c|ccc|cc|ccc}
\hline
\multirow{2}{*}{}      & \multirow{2}{*}{\textbf{Method}}    & \multicolumn{3}{c|}{\textbf{Feature Extraction}}                                & \multicolumn{2}{c|}{\textbf{Feature Matching}}                         & \multicolumn{3}{c}{\textbf{FIVR-200K ↑}}         \\
                       &                                     & \textbf{Dim} & \textbf{GLOPs ↓}                & \textbf{Storage ↓}           & \textbf{Type} & \textbf{Time ↓}                & \textbf{DSVR}  & \textbf{CSVR}  & \textbf{ISVR}  \\ \hline
\multirow{3}{*}{\rotatebox{90}{$Frame$}}  & TCA$_f$\cite{shao2021tca}           & 1024  & 34.2G                           & 2155M                        & model-based                            & 35.2s                          & 0.877          & 0.830          & 0.703          \\
                       & Visil$_v$\cite{kordopatis2019visil} & 3840  & 32.9G                           & 8079M                        & model-based                            & 463.8s                         & 0.892          & 0.841          & 0.702          \\
                       & VRL$_f$\cite{he2022vrl}             & 512   & -                               & 1078M                        & embedding-based                           & -                          & \textbf{0.900} & \textbf{0.858} & \textbf{0.709} \\ \hline
\multirow{3}{*}{\rotatebox{90}{$Clip$}}  & 3D-CSL-B\cite{3dcsl}                & 768   & 196.0G                          & 208M                         & embedding-based                            & 4.1s                           & 0.879          & 0.835          & 0.711          \\
                       & \textbf{DRCA-B-K4}                   & 768   & 129.0G\textcolor{blue}{(↓1.5x)} & 208M                         & embedding-based                            & 4.1s                           & \textbf{0.890} & \textbf{0.850} & \textbf{0.723} \\
                       & \textbf{DRCA-S-K3}                   & 384   & 31.0G \textcolor{blue}{(↓6.3x)} & 104M \textcolor{blue}{(↓2x)} & embedding-based                            & 3.1s \textcolor{blue}{(↓1.3x)} & 0.870          & 0.828          & 0.694          \\ \hline
\end{tabular}
}
\caption{In the comparison of video retrieval performance on the FIVR-200K dataset, our DRCA method achieves the best trade-off between efficiency and performance compared to state-of-the-art methods and the highest performance among clip-level methods.}
\label{tab: retrieval-four-metrics}
\vspace{-1.5em}
\end{table*}

\vspace{-1em}
\subsection{Differentiable Context-aware Compression Module}
\vspace{-0.5em}
As shown in Fig \ref{fig:overview}, to adaptively reduce tokens, we employ differentiable saliency ranking predictor for saliency assessment and compress non-saliency frames with saliency frame reference compressor.

\vspace{0.2em}
\label{krcm}
\noindent\textbf{Differentiable Saliency Ranking Predictor} We calculate the saliency scores $s \in \mathbb{N}^{T}$ of the $T$ frames using the low-level frame features(we use the third transformer block output). Specifically, the salient score prediction network consists of a 3D convolutional layer, an average pooling layer, and two linear layers. 

Given the saliency scores $s$, we formalize the frame saliency prediction as a ranking problem: $R(s) = y \in \mathbb{N}^{T}$, where $y$ is the sequence of indices of the ranked $s$. Based on the sorted indices $y$, we define the first K frame as saliency frames, while maintaining the rest as non-saliency frames. We denote the corresponding tokens as $t^{K} \in \mathbb{R}^{k\times M \times N \times C}$ and $t^{NK} \in \mathbb{R}^{(T-k)\times M \times N \times C}$. To get the optimal sequence $y$, we matrixize the $y$ as $Y = [\mathit{I}_{y_1}, \mathit{I}_{y_2}, \dots, \mathit{I}_{y_N}] \in \{0,1\}^{T \times T}$, and the sorted frame sequence can be denoted as $X^{R}=Y^TX$. This is a non-differentiable discrete convex optimization problem and can be parameterized by the input $S$ as follows:
\vspace{-0.5em}
\begin{equation}
   \operatorname*{arg\,max}_{Y\in \mathcal{C}} <Y, S^T>
   \label{equ:argmax}
\end{equation}
\vspace{-0.2em}
where, $S^T = s\mathit{1}^T$, and $\mathcal{C}$ is the convex polytope constrain set defined as:
\vspace{-0.5em}
\begin{equation}
\begin{aligned}
\mathcal{C} = \{ & Y \in \mathbb{R}^{T \times T}: Y_{t, t} \ge 0, \mathit{1}^TY = 1, Y\mathit{1}\le 1, \\
& \sum_{i\in [T]}{iY_{i,k}} < \sum_{j \in [T]}{jY_{j,k'}, \forall k<k'} \}
\label{constraint_set}
\end{aligned}
\end{equation}
\vspace{-0.2em}
In order to optimize the non-differentiable ranking operation with back-propagation, we resort to the perturbed maximum method \cite{berthet2020pertured}, taking the expectations with respect to the random perturbation as the smoothed versions of Eq.\ref{equ:argmax}:
\vspace{-0.5em}
\begin{equation}
    Y_\sigma = \mathbb{E}_Z[ \operatorname*{arg\,max}_{Y\in \mathcal{C}} <Y, S^T+\sigma Z>]
\end{equation}
\vspace{-0.2em}
where, $Z$ is a uniform Gaussian distribution noise vector and $\sigma$ is a hyper-parameter. We compute the expectation using an empirical mean with $n$ independent samples for $Z$ in practice. In our experiments, $\sigma $ is tuned as $0.05$ and $n$ is set as $500$. When backward, the Jacobin of the forward pass can be calculated as:
\vspace{-0.5em}
\begin{equation}
    J_sY = \mathbb{E}_Z[\operatorname*{arg\,max}_{Y\in \mathcal{C}} <Y, S^T + \sigma Z>Z^T/\sigma]
\end{equation}
\vspace{-0.2em}
which is a special case where $Z$ follows a normal distribution so that the ranking problem can be optimized with back-propagation.

\noindent\textbf{Saliency Frame Reference Compressor}. Instead of directly downsample the resolution of non-saliency frame tokens, which may lose some information non-existent in saliency frames, we propose the Saliency Frame Reference Compressor to compress the non-saliency frame tokens by referring to the task-related information in saliency frame tokens so that the valid information in non-saliency frames is retained and enhanced as much as possible, as shown in Figure \ref{fig:overview}(b). Specifically, the low-resolution non-saliency frame tokens are computed as:
\vspace{-0.5em}
\begin{equation}
t_{low}^{NK} =\text{Softmax}\left(\frac{Q_{low}^{NK}{K_{low}^{K}}^T}{\sqrt{C} }\right)V_{low}^{K}
\label{crdc_equation}
\end{equation}
\vspace{-0.5em}
\begin{align*}
Q_{low}^{NK} = D(w_a^{T}t^{NK}), K_{low}^{K} = D(w_b^{T}t^{K}), V_{low}^{K} = D(w_c^{T}t^{K})
\end{align*}
where, $w_a$, $w_b$, $w_c$ are the weights of the linear layers and $D(\bullet)$ is a spatial downsampling operation to get the low-resolution tokens $Q_{low}^{NK}, K_{low}^{K}, V_{low}^{K}\in \mathbb{R}^{\frac{M}{h} \times \frac{N}{h} \times C}$ with $h^2 \times$ compression ratio. 
Finally, a residual connection is adopted to get the final compressed non-saliency frame:
\begin{equation}
    t_{com}^{NK} = t_{low}^{NK} + D(t^{NK})
\label{residual_connection}
\end{equation}
Now we get the multi-resolutions sequence $X^{MR} = [t^{K},$ $t_{com}^{NK}]$, with less tokens than $X$. It is worth noting that the computation cost increase of DCCM is negligible compared to the cost reduction achieved by compressing non-saliency frames in subsequent layers.

\vspace{-1em}
\subsection{Resolution-Align Transformer Layer}
\label{ratlayer}

To effectively learn from multi-resolution sequences $X^{MR}$, the attention scores in the RAT Layer are computed separately for spatial and temporal. This allows it to compute spatial attention within each frame with self-attention, which is not affected by the multi-resolution issue. However, as a consequence, the non-saliency frames are computed at a lower resolution, resulting in a significant reduction in computational complexity by a factor of $h^4\times$ compared to the original $(T-k)$ non-saliency frames.

For temporal attention, it is necessary to align the resolution of saliency and non-saliency frame tokens. This is achieved by downsampling all saliency frame tokens to $t_{low}^{K}$, aligning to the resolution of $t_{com}^{NK}$ at first. Subsequently, the entire token sequence$X^{MR}= concat(t_{low}^{K}, t_{com}^{NK})$ is fed to compute the temporal self-attention map $att \in \mathcal{R}^{T\times \frac{M}{h} \times \frac{N}{h} \times C}$. Finally, the tokens are computed with residual connections as follows:

\vspace{-1em}
\begin{align}
    t_{i+1}^{K} &= t_{i}^{K} + \mathcal{U}(att) \\
    t_{i+1}^{NK} &= t_{i}^{NK} + att
\end{align}
\vspace{-0.2em}
where, $\mathcal{U}(\bullet)$ denotes an upsampling operation. In order to keep the details of saliency frames in the original resolution, the $att$ is upsampled to $Att \in \mathbb{R}^{T\times M \times N \times C}$ and add with the residual connection to restore the original high resolution. Besides, every frame token is attached with it's corresponding position embedding during the patch embedding, even though the order of tokens are rearranged when concatenating, the attention computation won't be affected.

\vspace{-0.5em}
\section{Experiments}
\vspace{-0.5em}
\subsection{Experimental Setting}
\vspace{-0.5em}
\noindent\textbf{Model Definition} We denote various versions of DRCA as DRCA-$N$-K$i$, where $N$ refers to the backbone and $i$ indicates the number of saliency frames. There are two versions of the backbone: base (B) and small (S). The embedding dimensions in the Transformer layers are 768 and 384 for the base and small backbones, respectively.

\noindent\textbf{Training Details} In the training stage, we first train a high-resolution model without compressing frames and subsequently finetune the proposed DRCA model with various compression settings for saliency frames. Furthermore, in the retrieval experiments, following 3D-CSL\cite{3dcsl}, we used Kinetics-400\cite{kay2017kinetics} as the training dataset for unsupervised similarity learning. If not specified, we use a resolution of 224x224 as the default. 



\noindent\textbf{Dataset and Evaluation} The Fine-grained Incident Video Retrieval (FIVR-200K/5K)\cite{kordopatis2019fivr} dataset are employed in NDVR task for large-scale retrieval assessments and ablation study. The FIVR-200K dataset comprises 100 queries, and the goal of the retrieval task is to locate the relevant video from a test video database consisting of 225,960 videos. FIVR-5K is its subset, containing 50 queries and 5000 testing videos. These datasets consist of three sub-tasks with different relevance level, and evaluate the performance using mAP. In addition, we evaluated the performance of Dynamic Video Classification on the Mini-Kinetics\cite{kay2017kinetics} dataset. The dataset comprises 200 action categories, containing 120k training videos and 9.8k testing videos. We report Top1 and Top5 accuracy as classification metrics.


\begin{table}[]
\centering
\tabcolsep=10pt
\scalebox{0.85}{%
\begin{tabular}{cc|ccc}
\hline
\multirow{2}{*}{\textbf{}} & \multirow{2}{*}{\textbf{Method}} & \multicolumn{3}{c}{\textbf{FIVR-200K}}           \\  
                           &                                  & \textbf{DVSR}           & \textbf{CSVR}          & \textbf{ISVR}           \\ \hline
\multirow{3}{*}{\rotatebox{90}{$Video$}}     & DML\cite{kordopatis2017dml}                              & 0.398          & 0.378          & 0.309          \\
                           & TCA$_c$\cite{shao2021tca}                            & 0.570          & 0.553          & 0.473          \\
                           & VVS\cite{jo2023vvs}                             & 0.711          & 0.689          & 0.590          \\ \hline
\multirow{6}{*}{\rotatebox{90}{$Frame$}}     & Hnip\cite{lin2017hnip}                        & 0.819          & 0.764          & 0.622          \\
                           & Jo $et\, al.$\cite{jo2022exploring}                        & 0.896          & 0.833          & 0.674          \\
                           & TCA$_f$\cite{shao2021tca}                           & 0.877          & 0.830          & 0.703          \\
                           & Visil$_v$\cite{kordopatis2019visil}                          & 0.892          & 0.841          & 0.702          \\
                           & VRL$_f$\cite{he2022vrl}                            & \textbf{0.900} & \textbf{0.858} & 0.709          \\ \hline
\multirow{3}{*}{\rotatebox{90}{$Clip$}}      & VRL$_c$\cite{he2022vrl}                            & 0.876          & 0.835          & 0.686          \\
                           & 3D-CSL-B\cite{3dcsl}                      & 0.879          & 0.835          & 0.711          \\
                           & \textbf{DRCA-B-K4}                & \textbf{0.890} & \textbf{0.850} & \textbf{0.723} \\ \hline
\end{tabular}
}
 \caption[]{Comparison with SOTA on FIVR-200K dataset. }
\label{tab:sota-fivr200k}
\vspace{-1.5em}
\end{table}

\vspace{-1em}
\subsection{Comparison to the State-of-the-Art}
\vspace{-0.5em}
\noindent\textbf{Comparison with SOTA on Video Retrieval} Comparing the efficiency and accuracy of different video retrieval methods is complicated, as it involves multiple aspects in two stages: feature extraction and matching. Hence, we conducted a comparative analysis with several SOTA methods based on comparable mAP.  As shown in Table \ref{tab: retrieval-four-metrics} and Table\ref{tab:sota-fivr200k}, our DRCA method achieves the best trade-off between efficiency and performance compared to SOTA methods and the highest performance among clip-level methods. Specifically, frame-level methods come with expensive storage space and retrieval time, making them expensive in the real world. In contrast, our DRCA achieves competitive mAP in all three tasks with significantly less storage space ($10\times-38\times$ less) and faster retrieval time ($8.6\times-113\times$ faster). Notably, compared to 3D-CSL-B\cite{3dcsl}, another clip-level approach, our DRCA-B-K4 outperforms it by 1.1\%-1.5\% in mAP while demanding 35\% fewer GFLOPs. The superior performance of our method makes it more cost-effective and practical for large-scale video retrieval.  To further reduce computation costs, we developed a lightweight version, DRCA-S-K3, which achieves a $6.3\times$ inference acceleration, $2\times$ reduction in storage space, and $1.3\times$ reduction in retrieval time. 



\begin{table}[]
\centering
\tabcolsep=10pt
\scalebox{0.85}{%
\begin{tabular}{cccc}
\hline
\multicolumn{1}{c|}{\textbf{Method}}              & \textbf{Print} & \textbf{GFLOPs} & \textbf{Top1} \\ \hline
\multicolumn{4}{l}{\colorbox{gray!20}{\textit{Multi-Network-Based Method}}}                                                        \\
 \multicolumn{1}{c|}{LiteEval\cite{liteeval}}      & NeurIPS19      & 99.0              & 61.0          \\
\multicolumn{1}{c|}{AR-Net\cite{arnet}}           & ECCV20         & 32.0              & 71.7          \\
\multicolumn{1}{c|}{VideoIQ\cite{sun2021dynamic}} & ICCV21         & 20.4            & 72.3          \\ \hline
\multicolumn{4}{l}{\colorbox{gray!20}{\textit{Skipping-Based Method}}}                                                           \\

\multicolumn{1}{c|}{AdaFocus\cite{adafocusplus}}  & ICCV21         & 38.6            & 72.9          \\
\multicolumn{1}{c|}{SMART\cite{gowda2021smart}}   & AAAI21         & 20.4            & 72.3          \\
\multicolumn{1}{c|}{OCSampler\cite{ocsampler}}    & CVPR22         & 21.6            & 73.7          \\
\multicolumn{1}{c|}{TSQNet\cite{xia2022temporal}} & ECCV22         & 19.7            & 73.2          \\
\multicolumn{1}{c|}{D-STEP\cite{raviv2022dstep}}  & BMVC22         & 12.4            & 65.4          \\ \hline
\multicolumn{4}{l}{\colorbox{gray!20}{\textit{Compression-Based Method (ours)}}}                                                        \\
\multicolumn{1}{c|}{\textbf{DRCA-S-K5(160x160)}}   & -              & 17.9            & \textbf{73.9} \\
\multicolumn{1}{c|}{\textbf{DRCA-S-K6(128x128)}}   & -              & \textbf{12.1}   & 71.9          \\ \hline
\end{tabular}
} 
\caption{Comparison with SOTA on Mini-Kinetics dataset. }
\label{tab:sota-classification}
\vspace{-2em}

\end{table}


\noindent\textbf{Comparison with SOTA on Video Classification}  
We compared several SOTA methods in dynamic video recognition on the Mini-Kinetics, including \textit{Multi-Network-Based} and \textit{Skipping-Based methods}. As shown in Table \ref{tab:sota-classification}, compared to the recent advances OCSampler\cite{ocsampler}, which utilizes temporal sampling, our method achieved \textbf{0.2\%} higher accuracy while reducing computational costs by \textbf{21\%}. Besides, compared to another advances D-STEP\cite{raviv2022dstep}, which utilizes sampling in both temporal and spatial and achieves the lowest computational cost, our approach outperformed it with a \textbf{6.5\%} higher accuracy while reducing the computational cost by \textbf{2.4\%}.


\vspace{-1em}
\subsection{Quantitative Analysis}
\vspace{-0.5em}
\label{ablation_study}

 \noindent\textbf{(1) The Effectiveness of DCCM}  To study DCCM in our method, we conduct a comparative analysis with two frame sampling strategies: i) Uniform: selecting saliency frames uniformly, and ii) DCCM without Differentiable Rank. Additionally, we explored the potential of well-designed compression modules and leveraging information from other frames to enhance learning. We compare the proposed compression module with two baseline and three advanced methods\cite{swin, he2016deep}. The model is DRCA-S-K4 with 4 saliency frame in our experiments, which retaining only 50\% of frames in high resolution. The results in Table \ref{tab: k-RCM} demonstrate a significant performance decrease of \textbf{1.3\%-2\% } for other compression modules. However, when employing DCCM, we achieved a \textbf{34\%} reduction in computational cost with only \textbf{0.3\%} decrease in accuracy.

\begin{table}[b]
\centering
\scalebox{0.85}{%
\begin{tabular}{c|c|ccc}
\hline
\textbf{Exp}                                                                   & \textbf{Method} & \textbf{GFLOPs} & \textbf{Top1} & \textbf{Top5} \\ \hline
baseline                                                                       & wo-compression  & 50.8            & 76.5          & 92.3          \\ \hline
\multirow{3}{*}{\begin{tabular}[c]{@{}c@{}}Sampling\\  Strategy\end{tabular}}     & \cellcolor{grey!10} DCCM (ours)            & \cellcolor{grey!10} \textbf{33.6}   & \cellcolor{grey!10} \textbf{76.2} & \cellcolor{grey!10} \textbf{92.1} \\
                                                                               & Uniform         & \textbf{33.2}   & 75.0          & 91.6          \\
                                                                               & wo.DiffRank       & 33.6            &   74.7            & 90.8           \\ \hline
\multirow{4}{*}{\begin{tabular}[c]{@{}c@{}}Compression\\  Module\end{tabular}} & \cellcolor{grey!10} DCCM (ours)            & \cellcolor{grey!10} \textbf{33.6}   & \cellcolor{grey!10} \textbf{76.2} & \cellcolor{grey!10} \textbf{92.1} \\
                                                                               & ConvNet2D          & \textbf{33.4}   & 74.5          & 91.1          \\
                                                                               & ConvNet3D          & 34.5            & 74.7          & 91.3          \\
                                                                               & PatchMerge\cite{swin}      & 33.6            & 74.9          & 91.7          \\
                                                                               & ResBlock2D\cite{he2016deep}      & 34.3            & 75.0          & 91.4   \\
                                                                              & ResBlock3D\cite{he2016deep}      & 37.1            & 75.2          & 91.8 \\

                                                                               \hline
\end{tabular}
}
\caption{Comparison of the components in DCCM.}
\label{tab: k-RCM}
\vspace{-0.5em}
\end{table}

\begin{figure}[t]
\centering
\includegraphics[width=0.8\linewidth]{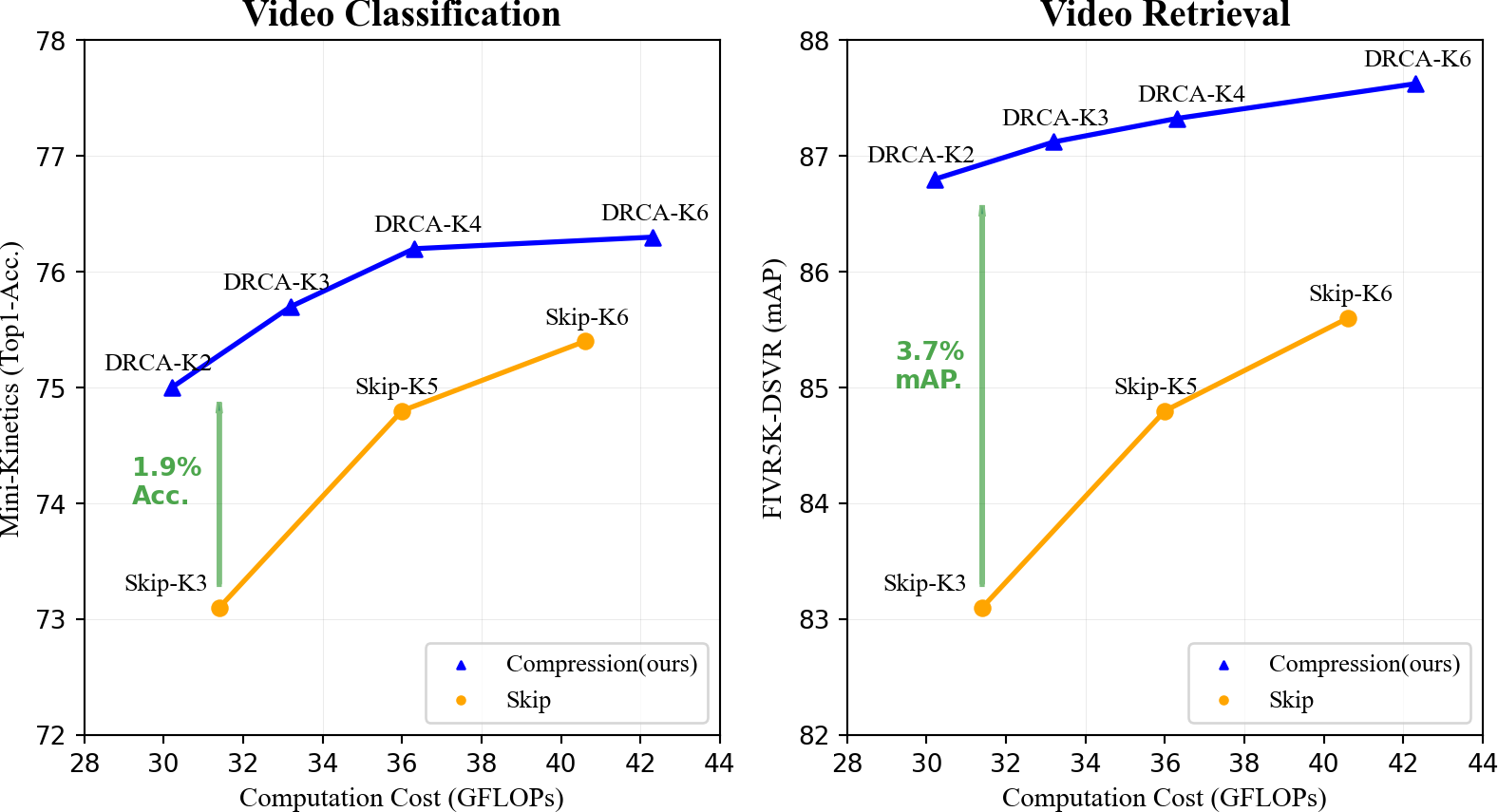}
\caption{The impact of resolution on video classification and retrieval. }
\label{figure:skip}
\vspace{-1.5em}
\end{figure}

\noindent\textbf{(2) The Effectiveness of RAT Layer} We explore the impact of different resolutions in two ways: (i) by varying the number of saliency frames and (ii) by comparing the proposed compression-based approach with non-saliency frame skipping mechanism. We use the standard Transformer to extract the high-resolution features when using the skipping mechanism. As shown in Fig \ref{figure:skip}, when considering the comparable GFLOPs, our method achieves higher accuracy in both classification and retrieval tasks, thus resulting in a better precision-efficiency trade-off. This is because the compression mechanism loses less information, which allows the model to use fewer saliency frames to reduce computation costs. Furthermore, the performance gap between compression and skipping is more significant in the retrieval task than in the classification task (\textbf{3.7\%} vs \textbf{1.9\%}), indicating that these two tasks have different requirements for input information. Specifically, the classification task only needs to focus on salient regions to make correct decisions. In contrast, the retrieval task requires retaining as much complete information as possible from the video to extract accurate representation.

\begin{figure}[h]
    \centering
  \includegraphics[width=0.8\linewidth]{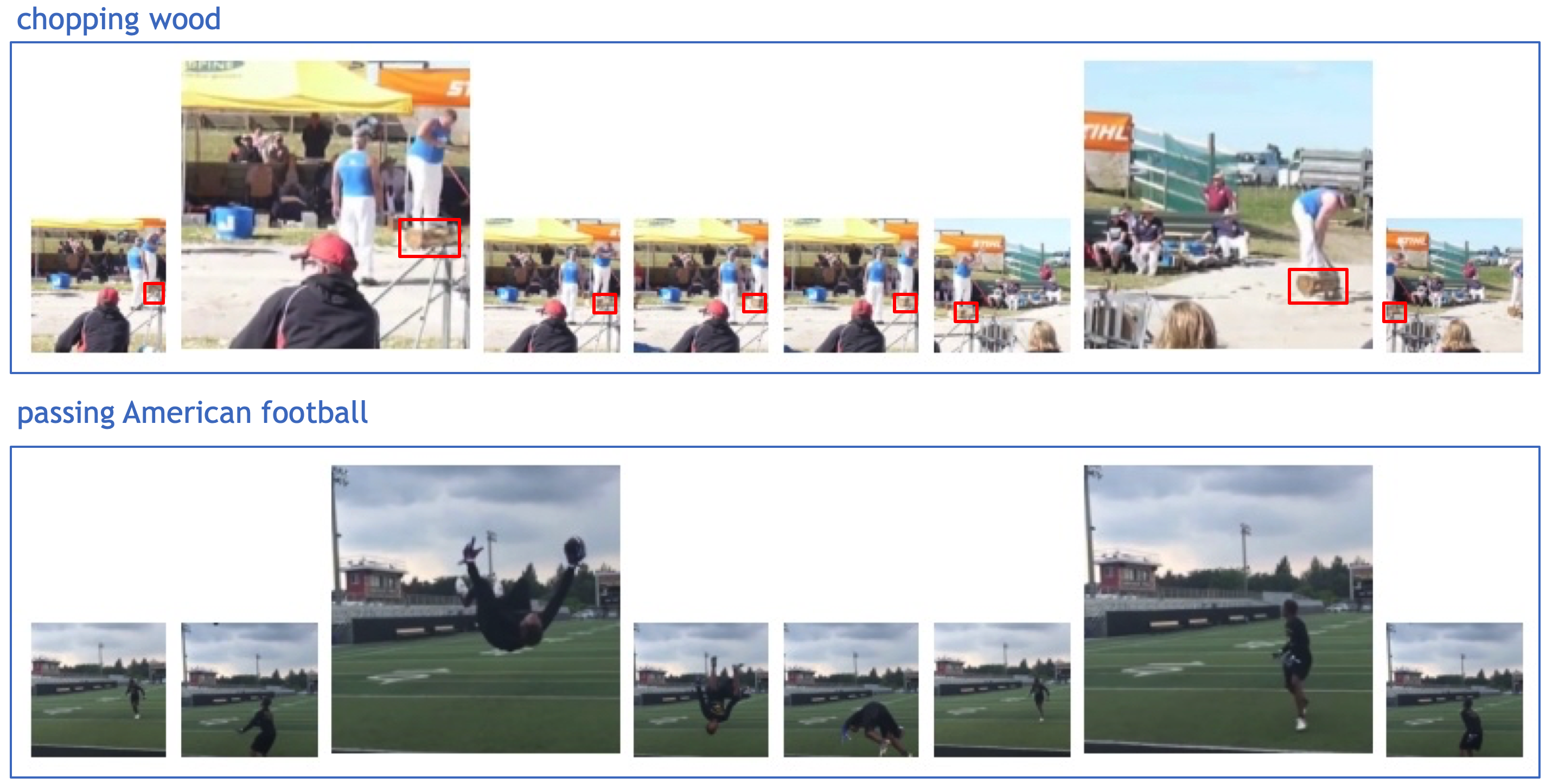}
    \caption{Examples of video classification. Our DRCA predicts 2 saliency frames from a uniformly sampled 8-frame video. }
    \label{fig:cls-example}
\vspace{-1.5em}
\end{figure}

\noindent\textbf{(3) Visualization Examples} As shown in Fig \ref{fig:cls-example}, our DRCA can select the most informative frames, such as in the case of \textit{chopping wood}, where it chooses the frames with the most salient wood (manually marked with red boxes for clarity). Furthermore, for some classes with fast-moving actions, such as \textit{passing a football}. These case studies suggest that the end-to-end optimized predictor can yield favorable results, enabling the model to capture relevant information even when decreasing input resolution. 


%

\vspace{-1em}

\vspace{-0.5em}
\section{Conclusions}
\vspace{-0.5em}
This paper propose a novel \textbf{Differentiable Resolution Compression and Alignment} network for efficient video representation learning, which improves computation efficiency while maintaining performance. The proposed DCCM can compress non-essential information with differentiable ranking, allowing for end-to-end optimization of the entire network. Furthermore, the proposed Resolution-Align Transformer Layer captures global temporal correlations in frame tokens with different resolutions while reducing spatial computation costs quadratically by utilizing fewer spatial tokens in low-resolution non-saliency frames. As a result, our method achieves the best trade-off between efficiency and performance on near-duplicate video retrieval and competitive results on dynamic video classification against SOTA methods.

\bibliographystyle{IEEEbib}
\bibliography{refs}

\begin{thebibliography}{10}

\bibitem{gowda2021smart}
Shreyank~N Gowda, Marcus Rohrbach, and Laura Sevilla-Lara,
\newblock ``Smart frame selection for action recognition,''
\newblock in {\em Proceedings of the AAAI Conference on Artificial Intelligence}, 2021, vol.~35, pp. 1451--1459.

\bibitem{adafocusplus}
Xiaopeng Zhang, Qiang Wang, Haoyu Liu, Xiaoyang Wang, and Dahua Lin,
\newblock ``Adafocus+: Adaptive focus for efficient video recognition,''
\newblock in {\em ICCV}, 2021.

\bibitem{ocsampler}
Bolun Cai, Han Wang, Xiangyu Zhang, and Dong Zhang,
\newblock ``Ocsampler: Compressing videos to one clip with single-step sampling,''
\newblock in {\em CVPR}, 2022.

\bibitem{xia2022temporal}
Boyang Xia, Zhihao Wang, Wenhao Wu, Haoran Wang, and Jungong Han,
\newblock ``Temporal saliency query network for efficient video recognition,''
\newblock in {\em Computer Vision--ECCV 2022: 17th European Conference, Tel Aviv, Israel, October 23--27, 2022, Proceedings, Part XXXIV}. Springer, 2022, pp. 741--759.

\bibitem{raviv2022dstep}
Avraham Raviv, Yonatan Dinai, Igor Drozdov, Niv Zehngut, Ishay Goldin, and Samsung Israel~R\&D Center,
\newblock ``D-step: Dynamic spatio-temporal pruning,''
\newblock 2022.

\bibitem{liteeval}
Chuang Liu, Ying Zhang, Ji~Lin, and Jianping Guo,
\newblock ``Liteeval: A coarse-to-fine framework for resource efficient video recognition,''
\newblock in {\em NIPS}, 2019.

\bibitem{arnet}
Yongxin Yang, Shuai Chen, Fanghui Liao, Peng Wang, Chen Qian, and Dong Zhang,
\newblock ``Ar-net: Adaptive frame resolution for efficient action recognition,''
\newblock in {\em ECCV}, 2020.

\bibitem{sun2021dynamic}
Ximeng Sun, Rameswar Panda, Chun-Fu~Richard Chen, Aude Oliva, Rogerio Feris, and Kate Saenko,
\newblock ``Dynamic network quantization for efficient video inference,''
\newblock in {\em Proceedings of the IEEE/CVF International Conference on Computer Vision}, 2021, pp. 7375--7385.

\bibitem{feichtenhofer2019slowfast}
Christoph Feichtenhofer, Haoqi Fan, Jitendra Malik, and Kaiming He,
\newblock ``Slowfast networks for video recognition,''
\newblock in {\em Proceedings of the IEEE/CVF international conference on computer vision}, 2019, pp. 6202--6211.

\bibitem{lin2019tsm}
J.~Lin, C.~Gan, and S.~Han,
\newblock ``Tsm: Temporal shift module for efficient video understanding,''
\newblock in {\em Proceedings of the IEEE/CVF international conference on computer vision}, 2019.

\bibitem{timesformer}
Gedas Bertasius, Heng Wang, and Lorenzo Torresani,
\newblock ``Is space-time attention all you need for video understanding?,''
\newblock in {\em ICML}, 2021, vol.~2, p.~4.

\bibitem{mvit}
X.~Liu, L.~Xu, T.~Zhang, J.~Yang, Z.~Wang, and H.~Li,
\newblock ``Mvit: Multiscale vision transformers for large-scale remote sensing image classification,''
\newblock in {\em CVPR}, 2021.

\bibitem{swin}
Ze~Liu, Yutong Lin, Yue Cao, Han Hu, Yixuan Wei, and Zheng Zhang,
\newblock ``Swin transformer: Hierarchical vision transformer using shifted windows,'' 2021.

\bibitem{rao2000reinforcement}
RPN Rao,
\newblock ``Reinforcement learning: An introduction,'' 2000.

\bibitem{gumblesoftmax}
Eric Jang, Shixiang Gu, and Ben Poole,
\newblock ``Categorical reparameterization with gumbel-softmax,''
\newblock {\em arXiv preprint arXiv:1611.01144}, 2016.

\bibitem{vit}
Alexey Dosovitskiy, Lucas Beyer, Alexander Kolesnikov, Dirk Weissenborn, Xiaohua Zhai, Thomas Unterthiner, Mostafa Dehghani, Matthias Minderer, Georg Heigold, Sylvain Gelly, et~al.,
\newblock ``An image is worth 16x16 words: Transformers for image recognition at scale,''
\newblock in {\em CVPR}, 2021.

\bibitem{shao2021tca}
Jie Shao, Xin Wen, Bingchen Zhao, and Xiangyang Xue,
\newblock ``Temporal context aggregation for video retrieval with contrastive learning,''
\newblock in {\em Proceedings of the IEEE/CVF Winter Conference on Applications of Computer Vision}, 2021.

\bibitem{kordopatis2019visil}
Giorgos Kordopatis-Zilos, Symeon Papadopoulos, Ioannis Patras, and Ioannis Kompatsiaris,
\newblock ``Visil: Fine-grained spatio-temporal video similarity learning,''
\newblock in {\em Proceedings of the IEEE/CVF International Conference on Computer Vision}, 2019.

\bibitem{he2022vrl}
Xiangteng He, Yulin Pan, Mingqian Tang, Yiliang Lv, and Yuxin Peng,
\newblock ``Learn from unlabeled videos for near-duplicate video retrieval,''
\newblock in {\em Proceedings of the 45th International ACM SIGIR Conference on Research and Development in Information Retrieval}, 2022, pp. 1002--1011.

\bibitem{3dcsl}
Rui Deng, Qian Wu, and Yuke Li,
\newblock ``3d-csl: self-supervised 3d context similarity learning for near-duplicate video retrieval,''
\newblock {\em arXiv preprint arXiv:2211.05352}, 2022.

\bibitem{berthet2020pertured}
Quentin Berthet, Mathieu Blondel, Olivier Teboul, Marco Cuturi, Jean-Philippe Vert, and Francis Bach,
\newblock ``Learning with differentiable pertubed optimizers,''
\newblock {\em Advances in neural information processing systems}, vol. 33, pp. 9508--9519, 2020.

\bibitem{kay2017kinetics}
Will Kay, Joao Carreira, Karen Simonyan, Brian Zhang, Chloe Hillier, Sudheendra Vijayanarasimhan, Fabio Viola, Tim Green, Trevor Back, Paul Natsev, et~al.,
\newblock ``The kinetics human action video dataset,''
\newblock {\em arXiv preprint arXiv:1705.06950}, 2017.

\bibitem{kordopatis2019fivr}
Giorgos Kordopatis-Zilos, Symeon Papadopoulos, Ioannis Patras, and Ioannis Kompatsiaris,
\newblock ``Fivr: Fine-grained incident video retrieval,''
\newblock {\em IEEE Transactions on Multimedia}, vol. 21, no. 10, pp. 2638--2652, 2019.

\bibitem{kordopatis2017dml}
G.~Kordopatis-Zilos, S.~Papadopoulos, I.~Patras, and Y.~Kompatsiaris,
\newblock ``Near-duplicate video retrieval with deep metric learning,''
\newblock in {\em Proceedings of the IEEE international conference on computer vision workshops}, 2017, pp. 347--356.

\bibitem{jo2023vvs}
Won Jo, Geuntaek Lim, Gwangjin Lee, Hyunwoo Kim, Byungsoo Ko, and Yukyung Choi,
\newblock ``Vvs: Video-to-video retrieval with irrelevant frame suppression,''
\newblock {\em arXiv preprint arXiv:2303.08906}, 2023.

\bibitem{lin2017hnip}
Jie Lin, Ling-Yu Duan, Shiqi Wang, Yan Bai, Yihang Lou, Vijay Chandrasekhar, Tiejun Huang, Alex Kot, and Wen Gao,
\newblock ``Hnip: Compact deep invariant representations for video matching, localization, and retrieval,''
\newblock {\em IEEE Transactions on Multimedia}, vol. 19, no. 9, pp. 1968--1983, 2017.

\bibitem{jo2022exploring}
Won Jo, Guentaek Lim, Joonsoo Kim, Joungil Yun, and Yukyung Choi,
\newblock ``Exploring the temporal cues to enhance video retrieval on standardized cdva,''
\newblock {\em IEEE Access}, vol. 10, pp. 38973--38981, 2022.

\bibitem{he2016deep}
Kaiming He, Xiangyu Zhang, Shaoqing Ren, and Jian Sun,
\newblock ``Deep residual learning for image recognition,''
\newblock in {\em Proceedings of the IEEE conference on computer vision and pattern recognition}, 2016, pp. 770--778.

\end{thebibliography}
\end{document}